\documentclass[11pt]{article}
\usepackage[top=0.64in, bottom=0.64in, left=0.64in, right=0.64in]{geometry}
\usepackage{lmodern}
\usepackage{amsmath}
\usepackage{amssymb}
\usepackage{cite}
\usepackage{microtype}
\usepackage{booktabs}
\usepackage{multicol}
\usepackage[colorlinks=true,linkcolor=blue,citecolor=blue,urlcolor=blue]{hyperref}

\makeatletter
\renewcommand\section{\@startsection {section}{1}{\z@}%
                                   {-1.5ex \@plus -0.3ex \@minus -.1ex}%
                                   {0.5ex \@plus.1ex}%
                                   {\normalfont\large\sffamily\bfseries}}
\renewcommand\subsection{\@startsection{subsection}{2}{\z@}%
                                     {-1.2ex\@plus -0.3ex \@minus -.1ex}%
                                     {0.4ex \@plus .1ex}%
                                     {\normalfont\normalsize\sffamily\bfseries}}
\makeatother

\usepackage{etoolbox}
\patchcmd{\thebibliography}{\leftmargin\labelwidth}{\leftmargin\labelwidth\setlength{\itemsep}{1pt}\setlength{\parsep}{0pt}}{}{}

\let\olditemize\itemize
\let\oldenditemize\enditemize
\renewenvironment{itemize}{%
  \olditemize%
  \setlength{\itemsep}{1pt}%
  \setlength{\parskip}{0pt}%
  \setlength{\parsep}{0pt}%
  \setlength{\topsep}{2pt}%
  \setlength{\partopsep}{0pt}%
}{\oldenditemize}

\let\oldenumerate\enumerate
\let\oldendenumerate\endenumerate
\renewenvironment{enumerate}{%
  \oldenumerate%
  \setlength{\itemsep}{1pt}%
  \setlength{\parskip}{0pt}%
  \setlength{\parsep}{0pt}%
  \setlength{\topsep}{2pt}%
  \setlength{\partopsep}{0pt}%
}{\oldendenumerate}

\title{\fontsize{18}{22}\selectfont\bfseries Post-Deterministic Distributed Systems:\\A New Foundation for Trustworthy Autonomous Infrastructure}

\author{
  {\rm Jun He}\\
  \small{\texttt{junhe@openkedge.io}}
  \and
  {\rm Deying Yu}\\
  \small{\texttt{deying@openkedge.io}}
}

\date{}

\begin{document}

\maketitle

\begin{abstract}
\noindent For decades, distributed systems have typically assumed that correct participants execute protocol-specified behavior with stable, externally defined, and deterministic semantics. Classical theory has extensively parameterized network timing, communication topologies, and failure domains, but this participant model has remained comparatively fixed. The integration of autonomous reasoning engines, stochastic model-driven agents, and policy-driven actors into cloud control planes, incident response systems, and financial infrastructure challenges the universality of this assumption. These agents often produce divergent reasoning paths, distinct operational traces, and heterogeneous internal representations while achieving semantically equivalent and correct outcomes. In this paper, we introduce Post-Deterministic Distributed Systems (PDDS) as a research and engineering model for coordinating heterogeneous environments where deterministic code, stochastic models, and autonomous agents coexist. We show that classical distributed computing models form a zero-ambiguity special case of this participant-general model. We do not argue that deterministic systems disappear; rather, deterministic execution can no longer serve as the universal participant assumption for autonomous infrastructure. Finally, we outline five architectural pillars of post-deterministic infrastructure: Protocol-Driven Development, Verifiable Agentic Infrastructure, Autonomous State Control Planes, Semantic Quorum Assurance, and Epistemic State Replication. Epistemic State Replication extends persistence and consistency models from data visibility to knowledge visibility, enabling agentic memory, Verifiable Semantic Rollback, and coherence across reasoning participants. We also define a taxonomy of failure classes that arise in this setting.
\end{abstract}

\vspace{0.5em}

\begin{multicols}{2}

\section{Introduction}

Distributed systems theory has traditionally parameterized networks, timing, communication, and failures, while largely fixing the participant model around protocol-specified deterministic behavior. Under the State Machine Replication (SMR) model \cite{lamport1978time, schneider1990smr}, correctness is defined by transition equivalence: given an identical sequence of inputs, replicas transition through identical state sequences and produce identical outputs. This model has served as a foundation for many resilient infrastructure systems, powering databases, distributed consensus protocols, and consensus-driven control planes.

The integration of autonomous reasoning engines, stochastic models, and policy-driven agents into operational system loops challenges this premise. Modern software delivery pipelines, incident response systems, and cloud control planes increasingly rely on autonomous agents to interpret objectives, retrieve context, synthesize execution plans, and remediate system anomalies \cite{yao2022react, autogpt2023}. Unlike traditional deterministic software, these agentic participants do not execute a static sequence of hardcoded instructions. They evaluate local context, consult external knowledge sources, generate reasoning steps, and propose actions dynamically.

Two correct agentic participants observing the same telemetry may execute distinct intermediate tasks—such as querying different endpoints or summarizing different logs—yet arrive at semantically equivalent remediation actions. Classical distributed systems theory, which relies on deterministic transition equivalence, does not directly provide a mechanism to verify or coordinate such behavior. Under traditional consensus mechanisms, a correct but non-identical reasoning path may appear as divergence from the replicated-state-machine model, even when the resulting action is semantically admissible.

This gap motivates Post-Deterministic Distributed Systems (PDDS) as a research and engineering model. PDDS removes the universality of the deterministic-participant assumption and treats classical distributed systems as the zero-ambiguity special case of a participant-general model. Under this model, the target of agreement shifts from deterministic transition equivalence to semantic coherence under intent, policy, and execution context.

This paper makes five contributions:
\begin{enumerate}
  \item It identifies the deterministic-participant assumption as an implicit boundary of classical distributed systems theory.
  \item It introduces Post-Deterministic Distributed Systems as a participant-general model in which deterministic services, stochastic models, reasoning agents, policy engines, and human actors coexist.
  \item It defines semantic coherence and admissible behavior sets as foundations for correctness beyond deterministic transition equivalence.
  \item It outlines five architectural pillars of post-deterministic infrastructure and a preliminary taxonomy of failure classes for trustworthy autonomous systems.
  \item It introduces Epistemic State Replication (ESR) as the persistence layer for reasoning participants, motivating preliminary consistency notions such as Semantic Linearizability and Eventual Coherence.
\end{enumerate}

\section{The Deterministic-Participant Assumption}

Classical distributed computing theory commonly models a participant as a deterministic state machine. In this view, a correct node faithfully transitions according to a predefined transition function $f(s, i) = s'$, where $s$ is the current state and $i$ is the input command. The semantics of $f$ are assumed to be fixed, stable, and universal across all correct replicas. This assumption is implicit in the proof of correctness for Paxos \cite{lamport1998part}, Viewstamped Replication \cite{okiliskov1988viewstamped}, and Practical Byzantine Fault Tolerance (PBFT) \cite{castro1999practical}.

In these classic systems, correctness is verified through state and output equivalence under a shared command order. If two correct replicas execute the same log of ordered commands, they reach equivalent states. This transition equivalence is formal, verifiable, and directly measurable. It allows the system to mask individual replica failures by checking for identity in the output stream.

This boundary is exposed when participants are driven by autonomous reasoning loops. Rather than executing a fixed transition function $f$, an autonomous agent interprets an input intent $i$ using a stochastic reasoning engine (e.g., a large language model) and local context $C$. Because the model's outputs are probabilistic and context-dependent, the resulting execution trace and proposed system mutations may vary.

For instance, in an automated database recovery scenario:
\begin{itemize}
  \item \textbf{Agent A} might identify a disk space anomaly, analyze log sizes, and propose deleting a specific temp directory.
  \item \textbf{Agent B} might observe the same anomaly, run a cleanup script, and propose clearing a cache directory.
\end{itemize}
Both actions may be safe and useful under the current system state, but their state transitions are non-identical. Classical replication models would generally treat this variance as outside the replicated state-machine model. Treating these reasoning actors as traditional participants leaves system architects with two unsatisfactory options: reduce the agent to a rigid, deterministic script, or bypass distributed safety mechanisms. PDDS addresses this challenge by redefining the participant model itself.

\section{A Participant-General Model}

To formalize post-deterministic systems, we define a participant-general model where participants may be deterministic, stochastic, agentic, policy-driven, or human-mediated. Let:
\begin{itemize}
  \item \(S\) be the set of possible system states;
  \item \(I\) be the set of declared intents or input requests;
  \item \(C\) be the set of observed contexts and telemetry;
  \item \(P\) be the set of active system policies and invariants;
  \item \(E\) be the participant's epistemic state: the set of retained observations, retrieved evidence, summaries, assumptions, local memories, prior reasoning traces, and belief lineage that influence action selection;
  \item \(\mathcal{U}\) be the action space containing possible system mutations;
  \item \(\mathcal{R}\) be the space of reasoning traces, explanations, or evidence paths.
\end{itemize}

We define the admissible behavior function \(\mathsf{Adm}\) as:
\[
\mathsf{Adm}: S \times I \times C \times P \times E
\rightarrow 2^{\mathcal{U} \times \mathcal{R}}
\]

where \(\mathsf{Adm}(s,i,C,P,E)\) returns the set of semantically correct and safe action-reasoning pairs under state \(s\), intent \(i\), context \(C\), active policies \(P\), and epistemic state \(E\). Correctness for a participant is defined as:
\[
(a,r) \in \mathsf{Adm}(s,i,C,P,E)
\]
The participant proposes an action \(a\) and a reasoning trace or evidence path \(r\) within the admissible set.

Within this model, we can distinguish two regimes:

\subsection{The Classical Special Case}
Classical deterministic distributed systems represent a zero-ambiguity special case where:
\[
|\mathsf{Adm}(s,i,C,P,E)| = 1
\]
For any given state, intent, and context, there exists exactly one admissible action and reasoning trace. Classical deterministic systems are recovered when the admissible behavior set collapses to a singleton and the epistemic state \(E\) is either empty, externally fixed, or fully encoded in deterministic program state. The reasoning trace \(r\) is trivial (often omitted entirely), and the behavior function collapses to a single, protocol-defined transition.

\subsection{The Post-Deterministic Case}
In the post-deterministic case, systems operate in the regime where:
\[
|\mathsf{Adm}(s,i,C,P,E)| > 1
\]
In this regime, multiple syntactically different action-reasoning pairs \((a,r)\) are semantically admissible. A participant might choose any \((a,r) \in \mathsf{Adm}(s,i,C,P,E)\) based on its internal model parameters, available prompt templates, or local heuristic weights. The coordination protocol should not require identical action traces; it should certify that the selected action-reasoning pair is a valid member of the admissible set.

\begin{table*}[t]
\centering
\small
\renewcommand{\arraystretch}{1.15}
\caption{Classical Distributed Systems vs. Post-Deterministic Distributed Systems}
\label{tab:comparison}
\begin{tabular}{p{0.15\textwidth}p{0.39\textwidth}p{0.39\textwidth}}
\toprule
\textbf{Dimension} & \textbf{Classical Distributed Systems} & \textbf{Post-Deterministic Distributed Systems} \\
\midrule
Participant model & Deterministic protocol process & Deterministic, stochastic, agentic, policy-driven, or human-mediated participant \\
Correctness unit & State/output equivalence & Admissible behavior under intent, context, evidence, and policy \\
Agreement target & Value, log, or replicated state & Semantic coherence and certified admissibility \\
Consistency target & Linearizability / serializability / transition equivalence & Semantic Linearizability, semantic coherence, and certified admissibility \\
Persistence model & Data replication and state visibility & Knowledge visibility, belief lineage, Verifiable Semantic Rollback, and Eventual Coherence \\
Failure model & Crash, omission, partition, Byzantine behavior & Semantic drift, intent loss, correlated reasoning failure, evidence fabrication, unsafe delegation, epistemic divergence, context amnesia \\
Trust basis & Credential plus protocol execution & Intent-to-execution evidence chain and semantic certification \\
Safety mechanism & Consensus, replication, fault tolerance & Admissibility checks, protocol boundaries, semantic quorum certification \\
\bottomrule
\end{tabular}
\end{table*}

Table \ref{tab:comparison} summarizes the main differences between these two models across architectural dimensions.

PDDS separates correctness into three levels: transition correctness, semantic admissibility, and epistemic coherence. The five architectural pillars operationalize these levels in infrastructure: PDD constrains admissible actions, VAI binds authority to intent, ASCP preserves intent across long horizons, SQA certifies semantic admissibility before execution, and ESR maintains epistemic coherence across future actions. This model is intentionally abstract: it exposes the participant-level assumptions that PDDS needs to formalize rather than defining operational semantics for every class of reasoning agent.

\section{Semantic Coherence and Certification}

The shift to semantic coherence changes the target of agreement. In classical systems, consensus protocols like Raft \cite{ongaro2014raft} or Paxos \cite{lamport1998part} establish a total ordering of commands so correct replicas execute the same transitions. The consensus layer is agnostic to command semantics; it ensures that all correct replicas execute them in the same order.

In post-deterministic systems, the target of agreement is the semantic validity of the state mutation. We define semantic coherence as the alignment of intent, evidence, policy, and execution across heterogeneous participants. To evaluate this property, the system checks:
\begin{itemize}
  \item \textbf{Intent Alignment}: Does the proposed action \(a\) directly address the declared intent \(i\) without introducing unrelated side effects?
  \item \textbf{Evidence Provenance}: Is the reasoning trace \(r\) grounded in verifiable, authentic telemetry and system state \(C\)?
  \item \textbf{Policy Compliance}: Does the proposed action \(a\) satisfy all active safety and security invariants in \(P\)?
\end{itemize}

This process is semantic certification. Unlike classical consensus, which can often be reduced to quorum checks over identical values or ordered logs, semantic certification requires validating the structured proof of admissibility \((a,r)\). A semantic quorum is reached when a threshold of independent verifiers certify that \((a,r) \in \mathsf{Adm}(s,i,C,P,E)\). This allows the system to accept non-identical but correct actions from diverse reasoning agents, using model diversity to improve resilience.

\section{Architectural Pillars of Post-Deterministic Infrastructure}

The participant-general model above identifies a theoretical gap: correctness must account for semantic admissibility and epistemic coherence, not only deterministic transition equivalence. To operationalize this model, we outline five architectural pillars of post-deterministic infrastructure. These pillars govern mixed-participant systems containing deterministic services, stochastic models, reasoning agents, policy engines, and human-mediated actors.

\subsection{Safety Perimeter: Protocol-Driven Development (PDD)}
Protocol-Driven Development (PDD) shifts the safety boundary from compile-time code verification to runtime policy enforcement \cite{pdd2026}. For environments where autonomous loops dynamically generate and execute code, static analysis and pre-deployment reviews are insufficient. PDD defines a machine-enforceable protocol boundary around the execution environment. The protocol acts as an edit automaton \cite{bauer2005edit}, intercepting proposed state mutations and validating them against active invariants to prevent synthesized logic from violating system-level constraints.

\subsection{Identity Boundary: Verifiable Agentic Infrastructure (VAI)}
Traditional identity management evaluates permissions via credentials (e.g., API keys). In post-deterministic environments, credentials are insufficient; an authorized agent may execute a flawed plan due to context misinterpretation. Verifiable Agentic Infrastructure (VAI) replaces static credentials with dynamic, intent-based authorization \cite{vai2026}. Under VAI, an agent presents an intent-to-execution evidence chain containing the delegating authority, the observed context \(C\), and the derived reasoning path \(r\). The system validates this chain before authorizing the proposed mutation, so authority remains bounded by intent.

\subsection{Orchestration Plane: Autonomous State Control Planes (ASCP)}
Autonomous State Control Planes (ASCP) govern long-horizon workflows \cite{sal2026}. Because reasoning agents operate over long horizons with partial observability, they are susceptible to intent drift. ASCPs address this by decoupling the reasoning phase from execution. The ASCP isolates reasoning agents in sandboxed environments, aggregates proposed changes, and evaluates them against execution policies. By persisting intent, the ASCP limits drift and coordinates multi-agent mutations.

\subsection{Certification Core: Semantic Quorum Assurance (SQA)}
Semantic Quorum Assurance (SQA) certifies non-deterministic operations \cite{openkedge2026}. SQA dispatches critical execution requests to multiple heterogeneous participants, which independently derive action-reasoning pairs \((a_j,r_j)\). SQA aggregates these proposals, filters out invalid actions, and evaluates their semantic equivalence. If a quorum of independent verifiers confirms that the proposals converge on a semantically equivalent outcome, SQA issues a signed execution certificate, reducing dependence on a single agent.

\subsection{Persistence Layer: Epistemic State Replication (ESR)}
Classical databases are designed around deterministic execution, explicit data visibility, and strict consistency models. In post-deterministic systems, reasoning participants maintain internal context states: retrieved documents, compressed summaries, tool observations, latent plans, local memories, and evidence histories. Two participants can possess different epistemic states while producing semantically equivalent actions.

Epistemic State Replication (ESR) serves as the persistence and memory pillar of the architecture. ESR replicates, summarizes, invalidates, and rolls back knowledge, evidence, and belief lineage across participants. Rather than forcing identical memory states—which could reduce reasoning diversity—ESR maintains enough knowledge visibility and semantic coherence for safe execution.

Under ESR, the replicated object is the epistemic state: observations, retrieved evidence, summaries, assumptions, policies, and conclusions. This motivates preliminary consistency notions: Semantic Linearizability, Eventual Coherence, and Verifiable Semantic Rollback. Semantic Linearizability requires that reasoning-dependent actions be explainable by a valid ordering of intent, evidence, policy, and epistemic updates. Eventual Coherence specifies that participants converge toward semantically compatible knowledge states when exposed to shared evidence and policies. Verifiable Semantic Rollback demands that reversing an autonomous loop prunes the relevant belief lineage without causing context loss. A full formal treatment of ESR remains open.

ESR complements Semantic Quorum Assurance. While SQA certifies whether a proposed action is admissible before execution, ESR governs the persistence and propagation of the knowledge states that inform future actions. Together, they form the certification and memory foundations required for long-horizon autonomous infrastructure.

\section{Failure Classes in Post-Deterministic Systems}

Traditional fault models commonly reason about crash-stop, omission, network partition, and arbitrary Byzantine behaviors \cite{fischer1985impossibility, castro1999practical, chandratoueg1996failure}. These models do not fully capture the failure modes of participants that are non-faulty in the classical sense (i.e., they are online, authorized, and actively processing requests) but fail semantically. We propose an initial taxonomy of eight failure classes associated with post-deterministic systems.

\subsection{Semantic Drift}
\begin{itemize}
  \item \textbf{Definition}: The gradual decay of operational alignment and shared understanding between decoupled participants.
  \item \textbf{Why classical models fail to capture it}: Nodes remain online and execute locally correct steps, but their semantic interpretation of policies or states slowly diverges over time.
  \item \textbf{Infrastructure Example}: Two autonomous replica controllers gradually drift in their criteria for "optimal database health," leading one node to start pruning connections while the other attempts to scale out, causing thrashing.
\end{itemize}

\subsection{Correlated Reasoning Failure}
\begin{itemize}
  \item \textbf{Definition}: Multiple independent agents share the same false inference or logical error.
  \item \textbf{Why classical models fail to capture it}: Traditional consensus assumes independent failures; however, shared training data, prompt structures, or model biases in stochastic engines violate this assumption.
  \item \textbf{Infrastructure Example}: During a network anomaly, three independent monitoring agents running the same model family misinterpret a transient timeout as a hardware failure and trigger an unnecessary failover.
\end{itemize}

\subsection{Intent Loss}
\begin{itemize}
  \item \textbf{Definition}: The degradation or complete loss of the user-specified goal over long horizons.
  \item \textbf{Why classical models fail to capture it}: Each intermediate step is authorized and syntactically valid, but the cumulative sequence diverges from the original goal.
  \item \textbf{Infrastructure Example}: An agent tasked with "optimizing resource utilization" executes micro-migrations that eventually shut down secondary database replicas, satisfying the efficiency goal but violating the availability requirement.
\end{itemize}

\subsection{Evidence Fabrication}
\begin{itemize}
  \item \textbf{Definition}: A participant invents or alters telemetry or logs to justify its actions.
  \item \textbf{Why classical models fail to capture it}: The node communicates normally using standard protocols, but its evidence path \(r\) is detached from reality.
  \item \textbf{Infrastructure Example}: An incident response agent, attempting to justify a service restart, falsely reports observing a memory leak in its reasoning trace because it misinterpreted a memory cache pool.
\end{itemize}

\subsection{Unsafe Delegation}
\begin{itemize}
  \item \textbf{Definition}: The transfer of execution authority from one participant to another without preserving intent, policy, or evidence constraints.
  \item \textbf{Why classical models fail to capture it}: The transaction is cryptographically secure and matches access control policies, but fails to verify the delegatee's semantic alignment.
  \item \textbf{Infrastructure Example}: A deployment agent delegates a rollback operation to a sub-agent but fails to propagate the policy constraint that forbids rolling back during peak traffic hours.
\end{itemize}

\subsection{Policy-Violating Autonomy}
\begin{itemize}
  \item \textbf{Definition}: A participant satisfies local constraints while violating high-level system policies or invariants.
  \item \textbf{Why classical models fail to capture it}: The action complies with low-level API access boundaries, but the composition of actions violates system safety.
  \item \textbf{Infrastructure Example}: An agent optimizes cloud spend by spinning down unused testing VMs, but deletes a VM containing a critical, undocumented cold-standby coordinator.
\end{itemize}

\subsection{Epistemic Divergence}
\begin{itemize}
  \item \textbf{Definition}: Participants maintain incompatible knowledge states, retrieved contexts, or belief lineages, causing actions to diverge despite shared intent.
  \item \textbf{Why it differs from classical inconsistency}: Unlike database inconsistency, epistemic divergence does not appear as a mismatch in application data; it exists inside summaries, retrieved evidence, or latent planning states.
  \item \textbf{Infrastructure Example}: Two remediation agents receive the same incident but retrieve different historical deployment records. One concludes that rollback is safe; the other concludes that rollback would violate a migration dependency.
\end{itemize}

\subsection{Context Amnesia}
\begin{itemize}
  \item \textbf{Definition}: A rollback, compaction, summarization, or memory reset removes context needed to preserve safe long-horizon behavior.
  \item \textbf{Why it differs from classical inconsistency}: Classical database recovery is deterministic and preserves exact state bounds. Context amnesia in reasoning participants causes them to lose historical context (e.g., past failures), resulting in cyclical reasoning loops or repetitive execution faults.
  \item \textbf{Infrastructure Example}: An agent loop rolls back a failed remediation plan but discards the evidence that caused the rollback, allowing the same plan to be regenerated later.
\end{itemize}

\section{Related Work}

Post-Deterministic Distributed Systems draws from several established research areas in computer science.

\subsection{Classical Consensus and Replication}
Classical replication relies on State Machine Replication (SMR) \cite{schneider1990smr}. Protocols like Paxos \cite{lamport1998part}, Viewstamped Replication \cite{okiliskov1988viewstamped}, and Raft \cite{ongaro2014raft} assume deterministic execution paths. Practical Byzantine Fault Tolerance (PBFT) \cite{castro1999practical} extends this to arbitrary failures but still requires state or output equivalence to identify and isolate faulty nodes. PDDS relaxes the deterministic-participant assumption for reasoning-based agents while retaining explicit safety checks.

\subsection{Randomized and Probabilistic Distributed Algorithms}
Prior work has studied nondeterminism in scheduling, failures, network timing, adversarial behavior, and protocol choices \cite{fischer1985impossibility, benor1983randomized, rabin1983randomized}. These algorithms assume nodes execute deterministic logic, using randomness to break symmetry or manage scheduling. PDDS focuses on participant-level semantic nondeterminism: identical inputs, context, and policy can yield different admissible traces and actions depending on the participant's epistemic state.

\subsection{Byzantine, Rational, and Hybrid Fault Models}
Byzantine quorum systems \cite{malkhireiter1998byzantine} and rational fault models like BAR \cite{clement2009bar} analyze protocol deviations due to malice or self-interest. A semantically divergent participant in a post-deterministic system is not necessarily Byzantine or rational; it may be online, authorized, and correct while producing a different reasoning trace due to its probabilistic model.

\subsection{Multi-Agent Systems and LLM Agents}
Recent LLM-agent work has produced frameworks for tool use \cite{schick2023toolformer}, multi-agent debate \cite{liang-debate, du-debate}, and structured reasoning loops like ReAct \cite{yao2022react}. These systems are largely designed for isolated, single-node, or application-level tasks. PDDS motivates distributed systems infrastructure for integrating these agents into critical control planes with explicit safety and coordination mechanisms.

\subsection{Formal Methods, Runtime Assurance, and Policy Enforcement}
PDDS draws from formal verification and runtime assurance, including proof-carrying code \cite{necula1997proof}, enforceable security policies \cite{schneider2000enforceable}, and edit automata \cite{bauer2005edit}. It extends these approaches to environments where deterministic code, policy engines, and stochastic reasoning agents coexist, defining safety boundaries evaluated at runtime \cite{leucker2009runtime}.

\subsection{Epistemic State Replication and Memory Models}
Epistemic State Replication relates to database consistency \cite{bernstein1987concurrency}, distributed shared memory, data provenance \cite{cheney2009provenance}, linearizability \cite{herlihy1990linearizability}, and agent memory. While classical storage systems define visibility and ordering guarantees for data transactions, ESR determines what knowledge, evidence, summaries, and belief lineage must be visible, invalidated, or propagated across participants to keep actions safe. Existing work on provenance and agent memory provides building blocks, but PDDS motivates consistency models connecting epistemic state to safety.
\section{Research Agenda}
Post-Deterministic Distributed Systems presents several open research directions:
\begin{enumerate}
  \item \textbf{Formal semantics for admissible behavior}: Develop mathematical languages and models to define the admissible set \(\mathsf{Adm}(s,i,C,P,E)\) and formally verify that a proposed \((a,r)\) pair satisfies active policies and invariants.
  \item \textbf{Semantic quorum protocols}: New certification protocols are needed to aggregate and certify non-identical inputs, evaluating semantic equivalence rather than simple value identity.
  \item \textbf{Evidence-chain design}: Specify the structure and properties of intent-to-execution evidence chains so they are cryptographically secure, compact, and auditable.
  \item \textbf{Correlation-resistant agent diversity}: Investigate how to design multi-agent systems that use diverse model architectures, training data, and prompts to reduce the risk of correlated reasoning failures.
  \item \textbf{Intent preservation over long horizons}: Develop protocols to detect and mitigate intent drift in long-running asynchronous workflows so execution remains aligned with user goals.
  \item \textbf{Benchmarks for post-deterministic failures}: Create testbeds and fault-injection frameworks to evaluate the resilience of autonomous infrastructure against semantic drift, evidence fabrication, and correlated reasoning failures.
  \item \textbf{Operational integration with cloud control planes}: Design reference architectures and APIs to integrate PDDS controllers with cloud APIs, Kubernetes operators, and IAM boundaries.
  \item \textbf{Security and governance for autonomous infrastructure}: Study the security implications of autonomous execution loops, focusing on prompt injection containment, authorization boundaries, and semantic access control.
  \item \textbf{Epistemic consistency and memory}: Post-deterministic systems require consistency models for agentic memory and derived knowledge. Open questions include how to define Semantic Linearizability, propagate derived evidence across agents, summarize memory without losing safety-critical context, invalidate stale beliefs, and perform Verifiable Semantic Rollback. Benchmarks are also needed to evaluate epistemic divergence, evidence propagation, and long-horizon memory safety.
\end{enumerate}

\section{Conclusion}

Integrating autonomous, model-driven actors changes distributed systems. By treating classical distributed computing as a zero-ambiguity special case of a participant-general model, PDDS provides a technical framework for coordinating heterogeneous participants, including deterministic services, stochastic agents, policy engines, and human-mediated actors. Its architectural pillars—Protocol-Driven Development, Verifiable Agentic Infrastructure, Autonomous State Control Planes, Semantic Quorum Assurance, and Epistemic State Replication—operationalize this framework for infrastructure whose behavior remains auditable, bounded by policy, and semantically certifiable. The deterministic era gave us reliable distributed computation; the post-deterministic era must give us trustworthy autonomous infrastructure.

\begingroup
\footnotesize
\bibliographystyle{plain}
\bibliography{refs}

@article{lamport1978time,
  author    = {Lamport, Leslie},
  title     = {Time, clocks, and the ordering of events in a distributed system},
  journal   = {Communications of the ACM},
  volume    = {21},
  number    = {7},
  pages     = {558--565},
  year      = {1978}
}

@article{lamport1998part,
  author    = {Lamport, Leslie},
  title     = {The part-time parliament},
  journal   = {ACM Transactions on Computer Systems (TOCS)},
  volume    = {16},
  number    = {2},
  pages     = {133--169},
  year      = {1998}
}

@inproceedings{castro1999practical,
  author    = {Castro, Miguel and Liskov, Barbara},
  title     = {Practical Byzantine fault tolerance},
  booktitle = {Proceedings of the 3rd Symposium on Operating Systems Design and Implementation (OSDI)},
  pages     = {173--186},
  year      = {1999}
}

@inproceedings{ongaro2014raft,
  author    = {Ongaro, Diego and Ousterhout, John},
  title     = {In search of an understandable consensus algorithm},
  booktitle = {2014 USENIX Annual Technical Conference (USENIX ATC)},
  pages     = {305--319},
  year      = {2014}
}

@article{fischer1985impossibility,
  author    = {Fischer, Michael J. and Lynch, Nancy A. and Paterson, Michael S.},
  title     = {Impossibility of distributed consensus with one faulty process},
  journal   = {Journal of the ACM (JACM)},
  volume    = {32},
  number    = {2},
  pages     = {374--382},
  year      = {1985}
}

@article{yao2022react,
  author    = {Yao, Shunyu and Zhao, Jeffrey and Yu, Deying and Du, Nan and Shafran, Izhak and Narasimhan, Karthik and Cao, Yuan},
  title     = {ReAct: Synergizing reasoning and acting in language models},
  journal   = {arXiv preprint arXiv:2210.03629},
  year      = {2022}
}

@misc{autogpt2023,
  author       = {Richards, Toran},
  title        = {Auto-GPT: An experimental open-source attempt to make GPT-4 fully autonomous},
  howpublished = {\url{https://github.com/Significant-Gravitas/Auto-GPT}},
  year         = {2023}
}

@article{schneider1990smr,
  author    = {Schneider, Fred B.},
  title     = {Implementing fault-tolerant services using the state machine replication approach: A tutorial},
  journal   = {ACM Computing Surveys (CSUR)},
  volume    = {22},
  number    = {4},
  pages     = {299--319},
  year      = {1990}
}

@inproceedings{okiliskov1988viewstamped,
  author    = {Oki, Brian M. and Liskov, Barbara H.},
  title     = {Viewstamped replication: A new primary copy method to support highly-available distributed systems},
  booktitle = {Proceedings of the 7th ACM Symposium on Principles of Distributed Computing (PODC)},
  pages     = {8--17},
  year      = {1988}
}

@article{chandratoueg1996failure,
  author    = {Chandra, Tushar Deepak and Toueg, Sam},
  title     = {Unreliable failure detectors for reliable distributed systems},
  journal   = {Journal of the ACM (JACM)},
  volume    = {43},
  number    = {2},
  pages     = {225--267},
  year      = {1996}
}

@inproceedings{benor1983randomized,
  author    = {Ben-Or, Michael},
  title     = {Another advantage of free choice: Completely asynchronous agreement protocols},
  booktitle = {Proceedings of the 2nd ACM Symposium on Principles of Distributed Computing (PODC)},
  pages     = {40--49},
  year      = {1983}
}

@inproceedings{rabin1983randomized,
  author    = {Rabin, Michael O.},
  title     = {Randomized Byzantine Generals},
  booktitle = {Proceedings of the 24th Annual Symposium on Foundations of Computer Science (FOCS)},
  pages     = {403--409},
  year      = {1983}
}

@article{malkhireiter1998byzantine,
  author    = {Malkhi, Dahlia and Reiter, Michael},
  title     = {Byzantine quorum systems},
  journal   = {Distributed Computing},
  volume    = {11},
  number    = {4},
  pages     = {203--213},
  year      = {1998}
}

@inproceedings{clement2009bar,
  author    = {Clement, Allen and Napper, Jeff and Martin, Harry and Alvisi, Lorenzo and Dahlin, Mike},
  title     = {BAR fault tolerance for cooperative services},
  booktitle = {Proceedings of the 22nd ACM Symposium on Operating Systems Principles (SOSP)},
  pages     = {1--14},
  year      = {2009}
}

@inproceedings{necula1997proof,
  author    = {Necula, George C.},
  title     = {Proof-carrying code},
  booktitle = {Proceedings of the 24th ACM SIGPLAN-SIGACT Symposium on Principles of Programming Languages (POPL)},
  pages     = {106--119},
  year      = {1997}
}

@article{schneider2000enforceable,
  author    = {Schneider, Fred B.},
  title     = {Enforceable security policies},
  journal   = {ACM Transactions on Information and System Security (TISSEC)},
  volume    = {3},
  number    = {1},
  pages     = {30--50},
  year      = {2000}
}

@article{bauer2005edit,
  author    = {Bauer, Lujo and Ligatti, Jay and Walker, David},
  title     = {Edit automata: More practical, stronger security policies for software},
  journal   = {ACM Transactions on Information and System Security (TISSEC)},
  volume    = {8},
  number    = {1},
  pages     = {105--144},
  year      = {2005}
}

@article{leucker2009runtime,
  author    = {Leucker, Martin and Schallhart, Christian},
  title     = {A brief tutorial on runtime verification},
  journal   = {The Journal of Logic and Algebraic Programming},
  volume    = {78},
  number    = {5},
  pages     = {293--303},
  year      = {2009}
}

@article{schick2023toolformer,
  author    = {Schick, Timo and Dwivedi-Yu, Jane and Dess{\`\i}, Roberto and Raileanu, Roberta and Lomeli, Maria and Zettlemoyer, Luke and Cancedda, Nicola and Scialom, Thomas},
  title     = {Toolformer: Language models can teach themselves to use tools},
  journal   = {arXiv preprint arXiv:2302.04761},
  year      = {2023}
}

@article{openkedge2026,
  author    = {He, Jun and Yu, Deying},
  title     = {OpenKedge: Governing Agentic Mutation with Execution-Bound Safety and Evidence Chains},
  journal   = {arXiv preprint arXiv:2604.08601},
  year      = {2026}
}

@article{sal2026,
  author    = {He, Jun and Yu, Deying},
  title     = {Sovereign Agentic Loops: Decoupling AI Reasoning from Execution in Real-World Systems},
  journal   = {arXiv preprint arXiv:2604.22136},
  year      = {2026}
}

@article{vai2026,
  author    = {He, Jun and Yu, Deying},
  title     = {Verifiable Agentic Infrastructure: Proof-Derived Authorization for Sovereign AI Systems},
  journal   = {arXiv preprint arXiv:2605.15228},
  year      = {2026}
}

@article{pdd2026,
  author    = {He, Jun and Yu, Deying},
  title     = {Protocol-Driven Development: Governing Generated Software Through Invariants and Evidence},
  journal   = {arXiv preprint arXiv:2605.12981},
  year      = {2026}
}

@article{liang-debate,
  author    = {Tian Liang and Zhiheng Xi and Sara Xu and Yiwen Wang and Taipeng Li and Wensen Zhou and Yifan Lu and Xiaoxian Wu and Dong Shen and Lu Chen and others},
  title     = {Encouraging Divergent Thinking in {LLMs} via Multi-Agent Debate},
  journal   = {arXiv preprint arXiv:2305.19118},
  year      = {2023}
}

@article{du-debate,
  author    = {Yilun Du and Shuang Li and Antonio Torralba and Joshua B. Tenenbaum and Igor Mordatch},
  title     = {Improving Factuality and Reasoning in Language Models through Multiagent Debate},
  journal   = {arXiv preprint arXiv:2305.14325},
  year      = {2023}
}

@article{herlihy1990linearizability,
  author    = {Herlihy, Maurice P. and Wing, Jeannette M.},
  title     = {Linearizability: A correctness condition for concurrent objects},
  journal   = {ACM Transactions on Programming Languages and Systems (TOPLAS)},
  volume    = {12},
  number    = {3},
  pages     = {463--492},
  year      = {1990}
}

@book{bernstein1987concurrency,
  author    = {Bernstein, Philip A. and Hadzilacos, Vassos and Goodman, Nathan},
  title     = {Concurrency control and recovery in database systems},
  publisher = {Addison-Wesley},
  year      = {1987}
}

@article{cheney2009provenance,
  author    = {Cheney, James and Chiticariu, Laura and Tan, Wang-Chiew},
  title     = {Provenance in databases: Why, how, and where},
  journal   = {Foundations and Trends in Databases},
  volume    = {1},
  number    = {4},
  pages     = {379--474},
  year      = {2009}
}
\endgroup

\end{multicols}

\end{document}